\documentclass{article}
\usepackage{spconf,amsmath,graphicx,hyperref}
\usepackage{cite}
\usepackage{amssymb,amsfonts}
\usepackage{algorithmic}
\usepackage{textcomp}
\usepackage{tabularx, xcolor, colortbl, booktabs, multirow, array} 
\newcolumntype{C}{>{\centering\arraybackslash}X} 
\usepackage[misc]{ifsym}
\usepackage{algorithm}
\usepackage[table]{xcolor}
\usepackage{makecell}
\usepackage{threeparttable}
\usepackage{pifont}
\usepackage{subcaption}  

\usepackage{CJKutf8}
\hypersetup{
    colorlinks=true,  
    linkcolor=red,    
    citecolor=green,  
    urlcolor=magenta  
}
\usepackage{fancyhdr}
\fancypagestyle{firstpage}{
  \fancyhf{}
  
  \fancyfoot[L]{\footnotesize
© 2026 IEEE. Personal use of this material is permitted. Permission from IEEE must be obtained for all other uses, in any current or future media, including reprinting/republishing this material for advertising or promotional purposes, creating new collective works, for resale or redistribution to servers or lists, or reuse of any copyrighted component of this work in other works.
  }
}

\title{Semantic Reformulation Entropy for Robust Hallucination Detection in QA Tasks}

\name{Chaodong Tong$^{\star,\dagger}$, 
Qi Zhang$^{\ddagger}$, 
Lei Jiang$^{\star}$, 
Yanbing Liu$^{\star,\dagger}$,
Nannan Sun$^{\star}$\thanks{Corresponding author: sunnannan@iie.ac.cn},
Wei Li$^{\ddagger}$}
\address{
$^{\star}$Institute of Information Engineering, Chinese Academy of Sciences, Beijing, China\\
$^{\dagger}$School of Cyber Security, University of Chinese Academy of Sciences, Beijing, China\\
$^{\ddagger}$China Industrial Control Systems Cyber Emergency Response Team, Beijing, China\\
\{tongchaodong, jianglei, liuyanbing, sunnannan\}@iie.ac.cn, \{bonniezhangqi,ai4cics\}@126.com
}

\begin{document}
%

\maketitle
\thispagestyle{firstpage}

\begin{abstract}
Reliable question answering with large language models (LLMs) is challenged by hallucinations, fluent but factually incorrect outputs arising from epistemic uncertainty. Existing entropy-based semantic-level uncertainty estimation methods are limited by sampling noise and unstable clustering of variable-length answers. We propose Semantic Reformulation Entropy (SRE), which improves uncertainty estimation in two ways. First, input-side semantic reformulations produce faithful paraphrases, expand the estimation space, and reduce biases from superficial decoder tendencies. Second, progressive, energy-based hybrid clustering stabilizes semantic grouping. Experiments on SQuAD and TriviaQA show that SRE outperforms strong baselines, providing more robust and generalizable hallucination detection. These results demonstrate that combining input diversification with multi-signal clustering substantially enhances semantic-level uncertainty estimation.
\end{abstract}


\section{Introduction}
\label{sec:intro}

Large Language Models (LLMs) excel at tasks such as question answering (QA), summarization, and dialogue \cite{tan2023can,zhou2024large}, but often produce fluent yet factually incorrect outputs, known as \emph{hallucinations} \cite{ji2023survey}. In QA, these are particularly problematic as users expect knowledge-grounded answers \cite{adlakha2024evaluating,farquhar2024detecting}, often arising from \emph{epistemic uncertainty} \cite{hullermeier2021aleatoric,abbasi2024believe}. This motivates leveraging model-internal uncertainty to detect hallucinations.

Existing detection methods can be broadly categorized by the type of uncertainty signal they exploit: (i) \emph{likelihood-based} signals, which rely on token-level probabilities \cite{liu2021token,zhang2023enhancing}; (ii) \emph{representation-based} signals, capturing hidden-state variability \cite{grewal2024improving}; (iii) \emph{prediction-based} signals, such as model-assigned truth probabilities $P(\text{True})$ \cite{farquhar2024detecting}; and (iv) \emph{semantic-level} signals, exemplified by Semantic Entropy (SE)~\cite{farquhar2024detecting}, which quantify meaning variability across multiple high-temperature sampled outputs.

\begin{figure}[t]
\centering
\includegraphics[width=\columnwidth]{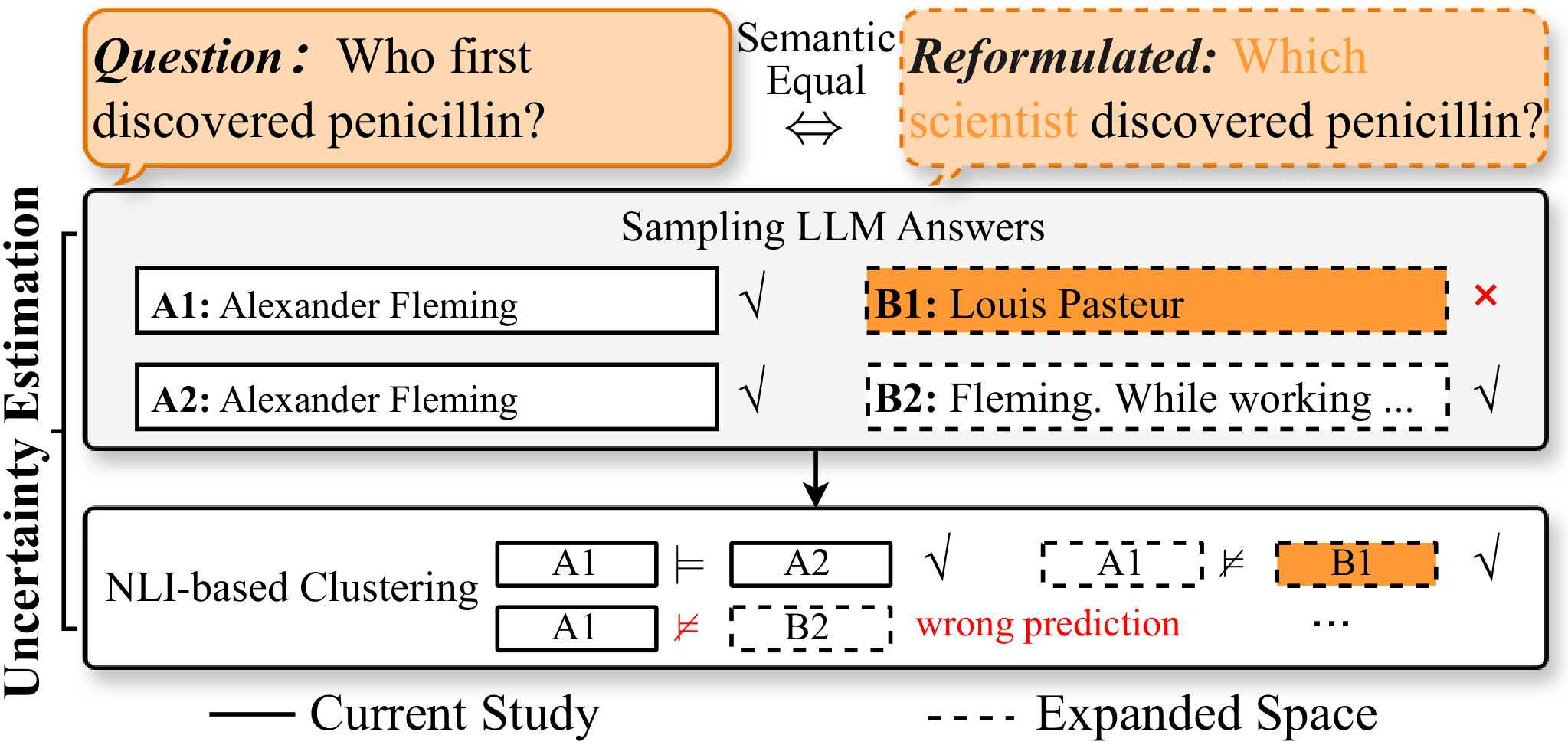}
\caption{Key limitations in QA uncertainty estimation: limited sampling space and fragile NLI-based clustering.}
\label{fig:overview}
\end{figure}

Semantic-level signals are promising, reflecting meaning variability in LLM outputs while requiring no internal model access. However, existing methods face two key limitations (Fig.~\ref{fig:overview}): (a) they rely on high-temperature sampling from a single input, making uncertainty estimates prone to biases from superficial decoder tendencies \cite{farquhar2024detecting,ma2025semantic}; and (b) clustering based solely on NLI is fragile, particularly for variable-length or semantically complex outputs \cite{arakelyan2024semantic}. Some recent works attempt to improve robustness using token-level uncertainty~\cite{ma2025semantic} or semantic perturbations~\cite{zhang2024vl}, but they do not systematically address these fundamental limitations.

In this work, we propose \textbf{Semantic Reformulation Entropy (SRE)}, a natural extension of semantic entropy aimed at more reliable estimation of epistemic uncertainty.
SRE enriches both the \emph{input} and \emph{output} sides of uncertainty estimation.
On the input side, \textbf{semantic reformulation (SR)} generates faithful paraphrases, expanding the estimation space and mitigating bias from superficial decoder patterns.
On the output side, \textbf{hybrid semantic clustering (HSC)} combines exact matches, embedding similarity, and bidirectional NLI via progressive, energy-based clustering, stabilizing assignments for variable-length or semantically complex outputs and supporting more reliable entropy estimation.
By combining SR and HSC, SRE provides a stable foundation for capturing genuine epistemic uncertainty at the semantic level in LLMs.

Our contributions are summarized as follows:
\begin{itemize}
\item We introduce SRE, which leverages input-side reformulation and a progressive, energy-based multi-signal clustering framewor named HSC to help separate epistemic uncertainty from superficial variability and produce robust semantic clusters.
\item We empirically show that SRE outperforms strong baselines on QA benchmarks (SQuAD, TriviaQA), providing more reliable and generalizable hallucination detection.
\end{itemize}


\section{Methodology}

\label{sec:method}
\subsection{Preliminaries}
We detect hallucinations via model-internal uncertainty. SE~\cite{farquhar2024detecting} measures output dispersion over clusters $C_\ell$:
\begin{equation}
\mathcal{H}_{SE}(x, M) = - \sum_{\ell=1}^{L} p(C_\ell \mid x, M) \log p(C_\ell \mid x, M),
\end{equation}
where $p(C_\ell \mid x, M)$ is the fraction or summed probability of outputs in $C_\ell$~\cite{farquhar2024detecting,kossen2024semantic}.  
We extend SE with \emph{SRE} (Fig.~\ref{fig:sre}), adding input reformulations and hybrid semantic clustering for more robust uncertainty estimation.

\subsection{Semantic Reformulation}
\begin{figure}[t] 
\centering \includegraphics[width=\columnwidth]{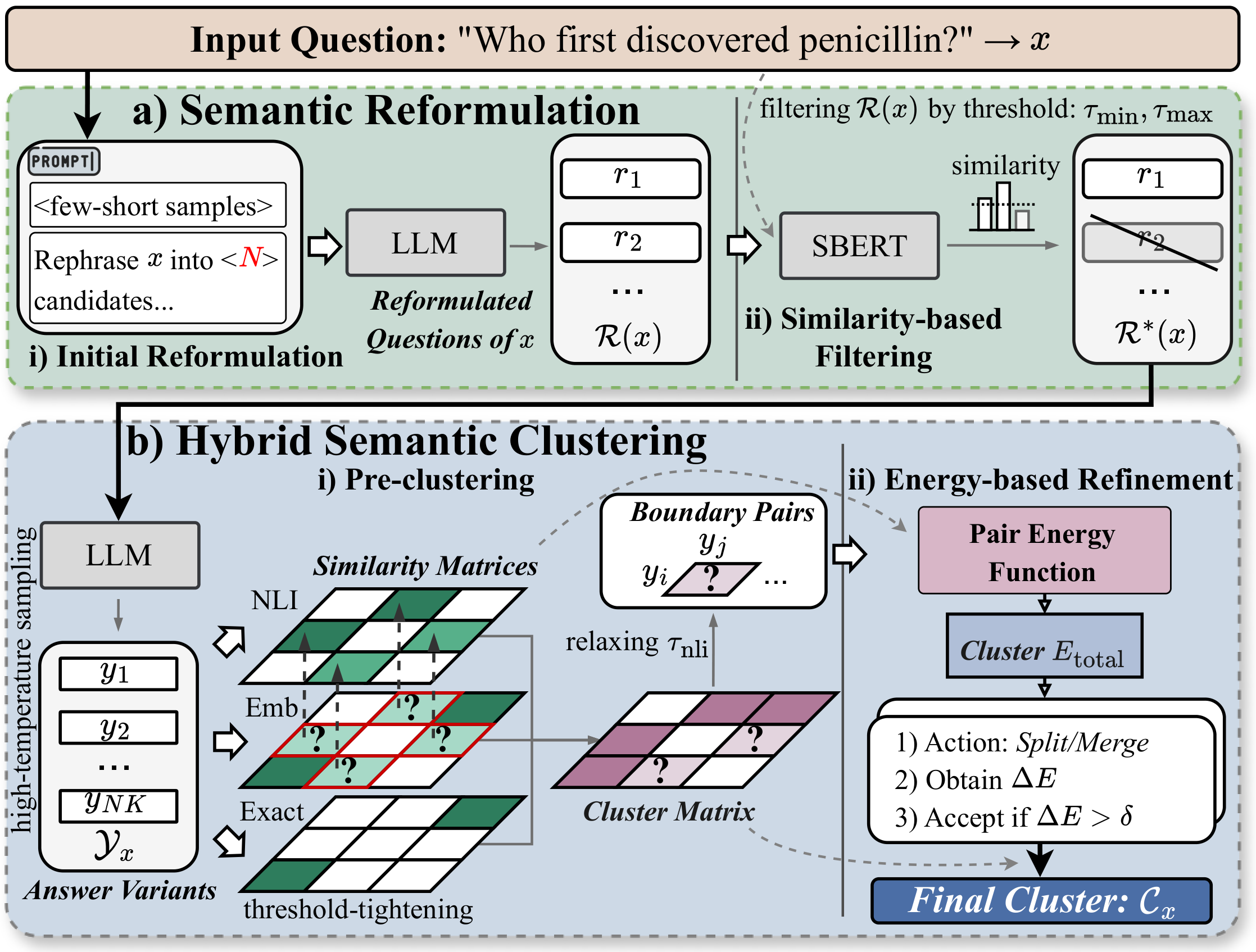} 
\caption{SRE: reformulating, sampling, and clustering.}
\label{fig:sre} 
\end{figure}

SE captures only output dispersion and is prone to superficial decoder biases. To address this, we generate \emph{semantic reformulations} $\mathcal{R}(x)=\{r_1,\dots,r_N\}$ via few-shot prompting, assessing model consistency across semantically equivalent queries while ensuring diversity and fidelity~\cite{zhang2024vl} (see Fig.~\ref{fig:sre}a). Each $r_i$ is scored against $x$ using cosine similarity:
\begin{equation}
s(r_i, x) = \frac{\langle e(r_i), e(x) \rangle}{\|e(r_i)\|\|e(x)\|},
\end{equation}
where $e(\cdot)$ is a sentence embedding~\cite{reimers2019sentencebert}. Candidates with $s(r_i, x) \notin [\tau_{\min},\tau_{\max}]$ or near-duplicates are removed, yielding $\mathcal{R}^*(x)$; if too few remain, additional reformulations are sampled to meet the desired count.

\subsection{Answer Sampling Process}
For each $r_i \in \mathcal{R}^*(x)$, outputs are sampled at high temperature with the prompt $\text{Prompt}(r_i) = \text{Few-shot examples} \,\|\, r_i$ to encourage diverse yet faithful responses.  

\textbf{Gold Labels.} Following~\cite{farquhar2024detecting}, gold labels are determined by comparing a single low-temperature output $y_i$ to the reference: $y_\text{gold} = 1$ if $y_i$ matches the reference, and $0$ otherwise. Multiple low-temperature samples yield consistent results, so we adopt this single-sample strategy.

\subsection{Hybrid Semantic Clustering}
SRE reliability depends on clustering quality. HSC clusters an input set $\mathcal{Y}_x$ of size $NK$ ($N$ reformulations, $K$ samples each) in two stages: pre-clustering and energy-based refinement (see Fig.~\ref{fig:sre}b).

\textbf{Pre-clustering (exact, embedding, NLI).}  
Outputs identical to each other (exact matches) are grouped directly. Remaining pairs with cosine similarity $sim(e(y_i),e(y_j))>\tau_\text{emb}$ are merged. For unmerged pairs, a \emph{DeBERTa-v2-xlarge-MNLI} model~\cite{he2021deberta} predicts entailment $p_{ij}^\text{ent}$ and contradiction $p_{ij}^\text{contra}$, and a merging score is defined:
\begin{align}
s_{ij}^\text{NLI} &= \lambda \frac{p_{ij}^\text{ent}+p_{ji}^\text{ent}}{2} 
\mathbf{1}\Big[\max(p_{ij}^\text{contra},p_{ji}^\text{contra})<\tau_\text{contra}\Big] \nonumber \\
&\quad + (1-\lambda) \max(p_{ij}^\text{ent},p_{ji}^\text{ent}),
\end{align}
merging pairs with $s_{ij}^\text{NLI}\ge\tau_\text{nli}$. Strict mode ($\lambda=1$) enforces contradiction filtering; loose mode ($\lambda=0$) uses only maximal entailment.

\textbf{Energy-based Boundary Refinement.}  
We refine \textit{boundary pairs}, i.e., pairs with entailment near $\tau_\text{nli}$ and contradiction below $\tau_\text{contra}$, to improve cluster structure.
The pair energy is defined as:
\begin{align}
E(i,j) &= 
\mathbf{1}_\text{same}(i,j) \Big[
1 - \big(\alpha\,\text{sim}(y_i,y_j) 
 + \beta\,\text{entail}(y_i,y_j)\big)
\Big] \notag\\
&\quad + \big(1-\mathbf{1}_\text{same}(i,j)\big)
\Big[1 - \gamma\,\text{contra}(y_i,y_j)\Big],
\end{align}
where $\mathbf{1}_\text{same}(i,j)=1$ if $y_i$ and $y_j$ are in the same cluster; $\text{entail}(\cdot,\cdot)$ and $\text{contra}(\cdot,\cdot)$ are NLI-based entailment and contradiction scores, and $\alpha,\beta,\gamma$ weight their contributions (in our experiments, $\alpha=0.3$, $\beta=0.7$, $\gamma=0.7$).

The total clustering energy is computed as the sum of average intra-cluster and inter-cluster energies:
\begin{equation}
E_\text{total} = \frac{1}{|\mathcal{P}_\text{intra}|} \sum_{(i,j)\in\mathcal{P}_\text{intra}} E(i,j) 
+ \frac{1}{|\mathcal{P}_\text{inter}|} \sum_{(i,j)\in\mathcal{P}_\text{inter}} E(i,j),
\end{equation}
where $\mathcal{P}_\text{intra}$ and $\mathcal{P}_\text{inter}$ denote the sets of intra-cluster and inter-cluster pairs, respectively.
For boundary pairs, we attempt split or merge operations and compute the total energy change:
\begin{equation}
\Delta E = E_\text{total}^\text{old} - E_\text{total}^\text{new},
\end{equation}
accepting the refinement if $\Delta E > \delta$, where we set $\delta = 0.1$.

This local greedy strategy efficiently corrects uncertain assignments while preserving confident clusters, ensuring stable and semantically coherent partitions.

\textbf{Progressive Integration and SRE Computation.}  
Outputs are clustered through a progressive integration of multiple signals, producing semantically coherent clusters $\mathcal{C}_x=\{C_1,\dots,C_m\}$. SRE is then defined as:
\begin{equation}
\mathcal{H}_\text{SRE}(x,M) = -\sum_{C\in \mathcal{C}_x} p_f(C) \log p_f(C),
\end{equation}
where $p_f(C)$ denotes the fraction of outputs in cluster $C$. By progressively refining clusters from coarse to fine, the method produces stable and robust clusters, providing a principled basis for entropy estimation.

\section{Experiments}
\subsection{Datasets}
We evaluate on SQuAD-v2~\cite{rajpurkar2018know} (contextualized QA) and TriviaQA~\cite{joshi2017triviaqa} (Table~\ref{tab:datasets}), sampling 500 validation instances each. To ensure representativeness, we compare context, question, and answer lengths, as well as the distribution of gold hallucination labels, with the full validation sets using Kolmogorov–Smirnov and Wasserstein distance tests~\cite{ramdas2017wasserstein}, and repeat sampling to verify stability.

\begin{table}[h]
\centering
\caption{Overview of QA datasets used in our experiments.}
\label{tab:datasets}
\begin{tabular}{lccc}
\toprule
Dataset & Context & \#Samples & Domain \\
\midrule
SQuAD-v2 & Full & 142,192 & Wikipedia \\
TriviaQA & Weak & 15,368 & Trivia \& Wikipedia \\
\bottomrule
\end{tabular}
\end{table}

\begin{table*}[ht]
\centering
\scriptsize
\setlength{\tabcolsep}{1.5pt} 
\renewcommand{\arraystretch}{1.0} 
\caption{Hallucination detection on SQuAD (with/without context) and TriviaQA (without context). \textbf{Bold}: best, \underline{underline}: second-best. Last row (HSC*) shows results without refinement and is excluded from ranking.}
\label{tab:main}
\resizebox{\textwidth}{!}{%
\begin{tabular}{l|ccc|ccc|ccc|ccc|ccc|ccc}
\toprule
\multirow{2}{*}{\textbf{Method}} 
& \multicolumn{6}{c|}{\textbf{SQuAD (context)}} 
& \multicolumn{6}{c|}{\textbf{SQuAD (no context)}} 
& \multicolumn{6}{c}{\textbf{TriviaQA (no context)}} \\
\cmidrule(lr){2-7} \cmidrule(lr){8-13} \cmidrule(lr){14-19}
& \multicolumn{3}{c}{Llama3-8B} & \multicolumn{3}{c|}{Qwen3-14B} 
& \multicolumn{3}{c}{Llama3-8B} & \multicolumn{3}{c|}{Qwen3-14B} 
& \multicolumn{3}{c}{Llama3-8B} & \multicolumn{3}{c}{Qwen3-14B} \\
\cmidrule(lr){2-4} \cmidrule(lr){5-7} \cmidrule(lr){8-10} \cmidrule(lr){11-13} \cmidrule(lr){14-16} \cmidrule(lr){17-19}
& AU & AR & F1@Best & AU & AR & F1@Best 
& AU & AR & F1@Best & AU & AR & F1@Best 
& AU & AR & F1@Best & AU & AR & F1@Best \\
\midrule
T-NLL           & 0.602 & 0.211 & 0.304 & 0.593 & 0.116 & 0.233 & 0.654& 0.861& 0.887& 0.558& 0.753& 0.837& 0.622& 0.390& 0.460& 0.595 & 0.478 & 0.561 \\
TE              & 0.256 & 0.069 & 0.276 & 0.291 & 0.036 & 0.139 & 0.235 & 0.625& 0.885 & 0.297& 0.593& 0.837& 0.172& 0.128& 0.457& 0.175 & 0.194 & 0.561 \\
TE (SR)         & 0.275 & 0.069 & 0.280 & 0.320 & 0.040 & 0.142 & 0.263 & 0.644 & 0.885 & 0.274 & 0.554& 0.837 & 0.186 & 0.124 & 0.457 & 0.171 & 0.178 & 0.561 \\
EV              & \textbf{0.755} & \underline{0.264} & \textbf{0.443} & 0.702 & \underline{0.134} & \textbf{0.330} & 0.728& 0.883 & \textbf{0.902} & 0.634& 0.781 & 0.833 & 0.838 & 0.535 & 0.704 & 0.807 & 0.613 & 0.719 \\
EV (SR)         & \underline{0.733} & 0.255 & 0.400 & \textbf{0.746} & 0.121 & 0.271 & 0.709 & 0.881 & 0.896 & 0.649 & 0.788 & 0.835 & 0.842 & 0.526& 0.667& 0.780 & 0.568 & 0.689 \\
$P(\text{True})$ & 0.576 & 0.195 & 0.309 & 0.650 & 0.102 & 0.206 & 0.703 & 0.880 & 0.890 & \underline{0.719}& \underline{0.843}& \underline{0.838}& 0.864 & \underline{0.560} & 0.701 & \underline{0.849} & \underline{0.662} & 0.702 \\
SE              & 0.614 & 0.222 & 0.300 & 0.533 & 0.089 & 0.142 & 0.766& 0.888& 0.885& 0.669& 0.821& 0.837 & 0.828 & 0.528 & 0.681 & 0.817 & 0.627 & 0.700 \\
SE (HSC)          & 0.730 & \textbf{0.276} & \underline{0.414} & \underline{0.706} & \textbf{0.135} & \underline{0.311} & \underline{0.801}& \underline{0.914}& \underline{0.900}& 0.701& 0.839& 0.836& \underline{0.870} & 0.557 & \underline{0.711} & 0.833 & 0.653 & 0.721 \\
\midrule
SRE (SR)        & 0.547 & 0.181 & 0.279 & 0.575 & 0.097 & 0.193 & 0.751& 0.884& 0.887& 0.633& 0.807& 0.835& 0.840 & 0.526& 0.669 & 0.829 & 0.645 & \underline{0.726} \\
SRE (SR+HSC)    & 0.711 & 0.244 & 0.411 & 0.705 & 0.120 & 0.271 & \textbf{0.806}& \textbf{0.917}& \underline{0.900} & \textbf{0.754} & \textbf{0.851} & \textbf{0.840} & \textbf{0.871} & \textbf{0.566} & \textbf{0.730} & \textbf{0.887} & \textbf{0.685} & \textbf{0.743} \\
\midrule\midrule
SRE (SR+HSC*)    & 0.701& 0.232 & 0.407 & 0.688 & 0.114 & 0.265 & 0.787& 0.903& 0.891 & 0.739 & 0.840 & 0.838 & 0.849 & 0.533 & 0.683 & 0.882 & 0.674 & 0.738 \\
\bottomrule
\end{tabular}%
}
\end{table*}

\subsection{Implementation Details}
We implement SRE with Llama3-8B-Instruct and Qwen3-14B on 8×RTX 3090 GPUs. SR generates $N=3$ paraphrases ($\tau_{\min}=0.6$, $\tau_{\max}=0.95$), sampling $K=8$ outputs ($T=0.8$) with up to 5 few-shot examples, while HSC combines exact match, embedding similarity ($\tau_\mathrm{emb}=0.92$), and NLI entailment ($\tau_\mathrm{nli}=0.8$); sensitivity of core parameters in SRE is analyzed in Section~\ref{psa}.

\subsection{Baselines and Evaluation Metrics}
We compare our method against representative baselines: likelihood-based T-NLL~\cite{zhang2023enhancing} and token-level entropy TE~\cite{holtzman2020curious}, embedding-based EV~\cite{grewal2024improving}, prediction-based $P(\text{True})$~\cite{farquhar2024detecting}, and semantic-level SE~\cite{farquhar2024detecting}. Performance is evaluated using AUROC (AU), AURAC (AR)~\cite{farquhar2024detecting}, and F1@best; hallucinations are positive, and our AURAC rejects low-confidence (likely non-hallucination) samples and measures hallucination accuracy among the rest, unlike standard AURAC, making it more meaningful in context-rich, low-hallucination settings.

\subsection{Main Experiments and Ablation Studies}
We evaluate on SQuAD-v2~\cite{rajpurkar2018know} and TriviaQA~\cite{joshi2017triviaqa} under open-domain QA (no context) and extractive QA (with context) (Table~\ref{tab:main}). In open-domain QA, SRE outperforms baselines, achieving a maximum AUROC of 0.887 (+4\% over the strongest baseline) and higher F1@Best. In extractive QA, SRE surpasses standard semantic entropy, though low AUROC reflects rare hallucinations. Performance is consistent across datasets and model variants.

\begin{figure}[H]
    \centering
    \includegraphics[width=\columnwidth]{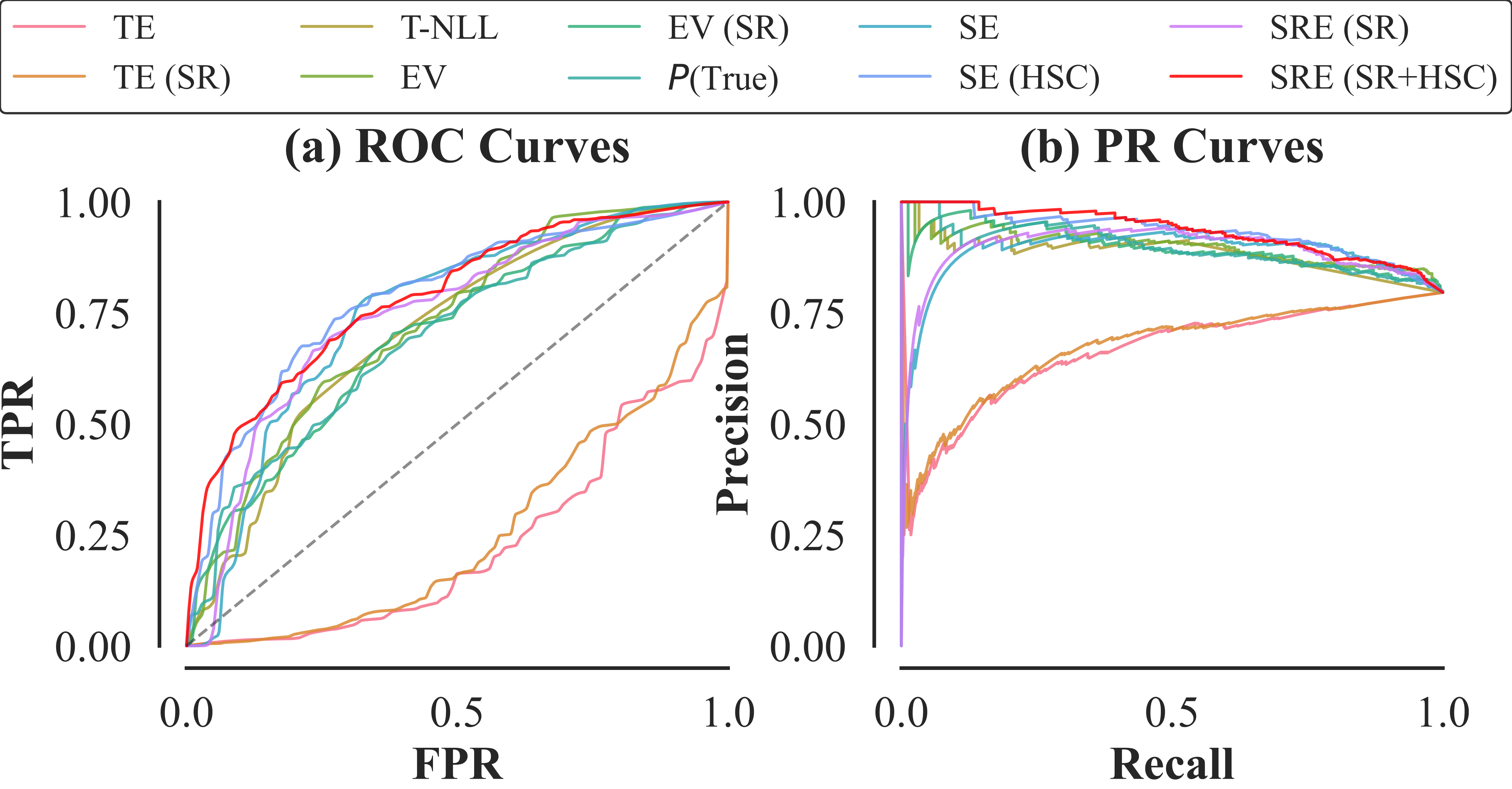}
    \caption{ROC (a) and PR (b) curves of all methods on SQuAD (no context) with Llama3-8B.}
    \label{fig:roc_pr}
\end{figure}

Ablations show that SR provides modest gains, while HSC, including boundary pair refinement, drives most improvements (up to +16\% AUROC). The refinement alleviates boundary ambiguities in coarse clusters; as it is greedy, the gains are moderate. ROC and PR curves (Fig.~\ref{fig:roc_pr}) indicate SRE achieves clearer separation between hallucinated and faithful samples, providing a more nuanced uncertainty assessment. Overall, SR combined with HSC effectively captures model uncertainty across datasets and QA settings.

\subsection{Parameter Sensitivity Analysis}
\label{psa}
\begin{figure}[t]
    \centering
    \includegraphics[width=\columnwidth]{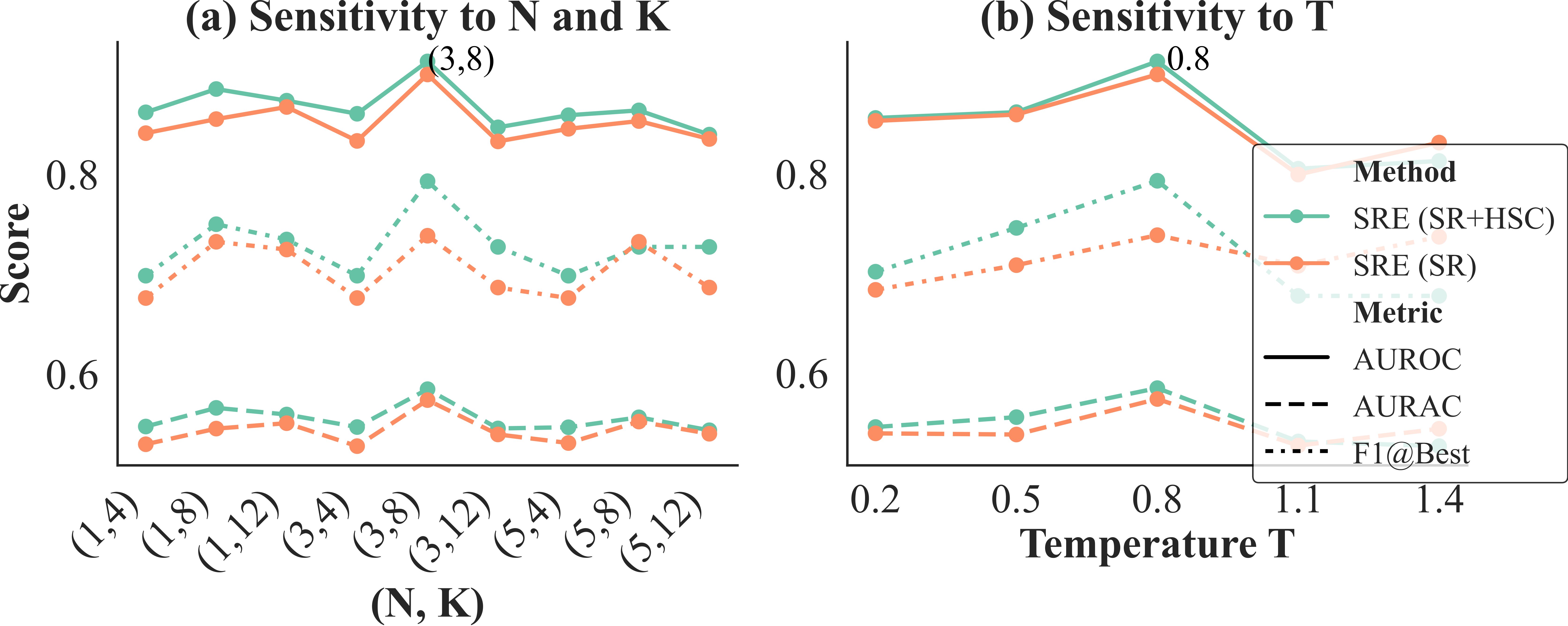}
    \caption{Parameter sensitivity analysis of SRE.}
    \label{fig:param_sensi}
\end{figure}

We evaluate SRE on 500 TriviaQA samples with Llama3-8B (Fig.~\ref{fig:param_sensi}). Performance depends on the number of reformulations ($N$), samples per reformulation ($K$), and sampling temperature ($T$). Best results are obtained with $N=3$, $K=8$, and $T=0.8$, while larger or extreme values degrade performance due to redundancy or lower-quality outputs. These results emphasize the need to balance diversity and quality for reliable uncertainty estimation.

\begin{figure}[!htb]
    \centering
    \begin{subfigure}[t]{\columnwidth}
        \centering
        \includegraphics[width=\columnwidth]{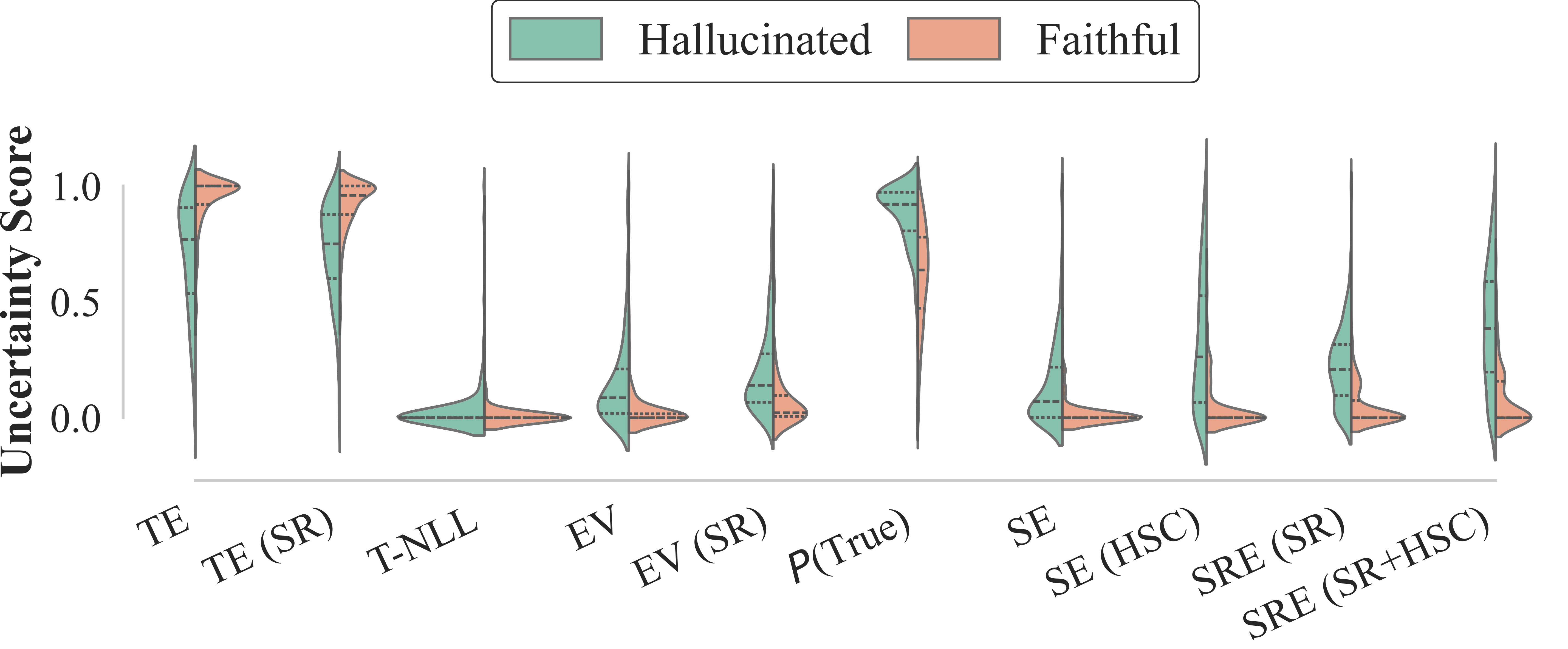}
        \caption{Uncertainty distributions of hallucinated vs. faithful samples for SRE and baselines.}
        \label{fig:violin}
    \end{subfigure}

    \vspace{0.5em} 

    \begin{subfigure}[t]{\columnwidth}
        \centering
        \includegraphics[width=\columnwidth]{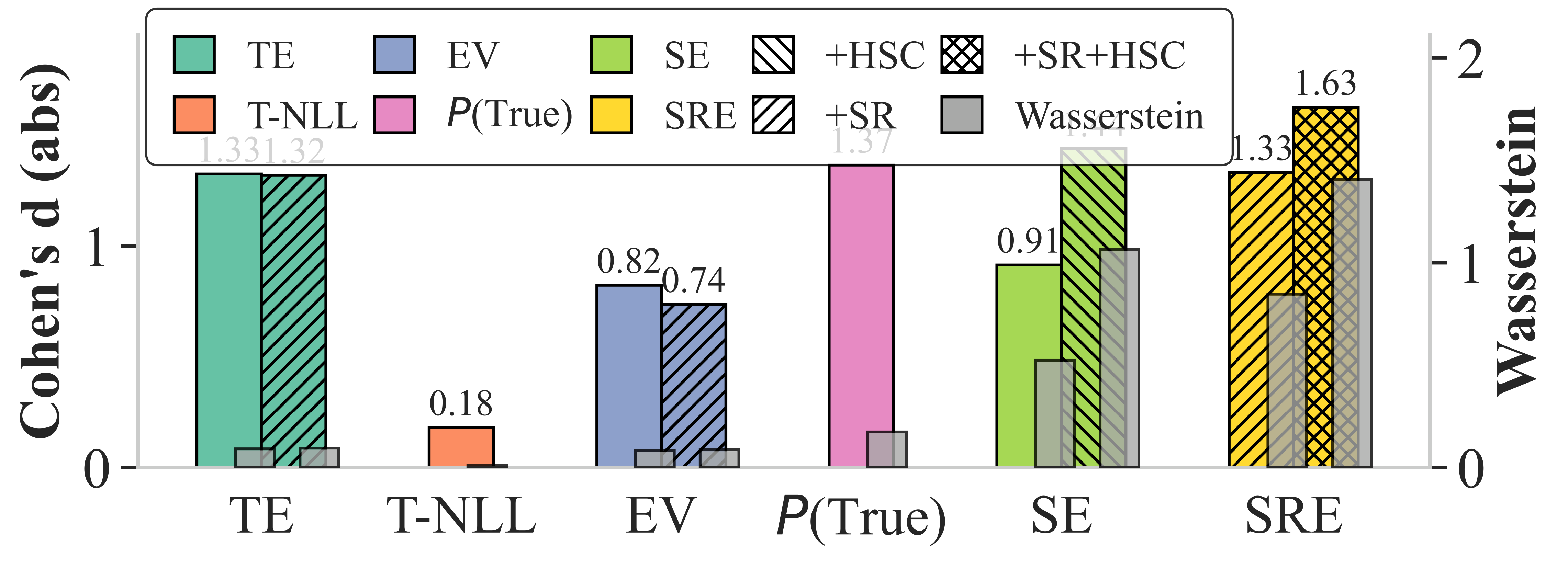}
        \caption{Cohen's $d$ and Wasserstein distance comparisons across methods.}
        \label{fig:coh}
    \end{subfigure}

    \caption{Distribution differences across methods.}
    \label{fig:combined}
\end{figure}

\subsection{Effect on Uncertainty Estimation and Score Distributions}
On SQuAD-v2 without context, the distributions of uncertainty scores in the violin (Fig.~\ref{fig:violin}) and bar plots (Fig.~\ref{fig:coh}) show that SRE effectively differentiates hallucinated from faithful samples. Compared to baselines (TE, T-NLL, EV, $P(\text{True})$, SE), SRE (SR+HSC) achieves a clearer separation, with hallucinated samples exhibiting a wider spread of uncertainty. Quantitative metrics, including Cohen's $d$ and Wasserstein distance, confirm that HSC drives most of the improvement while SR contributes moderately, whereas baselines display overlapping distributions and low distances. These results underscore SRE’s advantage and the crucial role of HSC in structuring the estimation space for reliable uncertainty measurement.

\section{Conclusion}
\label{sec:conclusion}
We propose SRE, leveraging input-side semantic reformulations and hybrid clustering to extend uncertainty estimation and more reliably capture epistemic uncertainty. Experiments on SQuAD and TriviaQA show that SRE outperforms strong baselines in hallucination detection. Consistent with related work, its efficiency is influenced by LLM sampling and NLI computation, which could be further optimized in a more systematic study. Nonetheless, our results suggest that expanding the uncertainty space and combining it with more precise estimation methods is a promising direction for enhancing LLM reliability.

\section{Acknowledgment}
This research is supported by the National Key R\&D Program of China (No. 2023YFC3303800).

\bibliographystyle{IEEEbib}
\bibliography{my}

\end{document}